\documentclass[11pt,letterpaper]{article}
\usepackage{algorithm}
\usepackage{algorithmic}
\usepackage{amsmath}
\usepackage{amsthm}
\usepackage{amssymb}
\usepackage{pifont}
\usepackage{graphicx}
\usepackage{xspace}
\usepackage{tabularx}
\usepackage{multicol}
\usepackage{multirow}
\usepackage{url}
\usepackage{wrapfig}
\usepackage{comment}
\usepackage{listings}
\usepackage{fancyvrb}
\usepackage{booktabs}
\usepackage{color,soul}
\usepackage[normalem]{ulem}

\usepackage[utf8]{inputenc}
\usepackage[ukrainian,bulgarian,russian,english]{babel}
\usepackage{times}
\usepackage{mathptmx}

\usepackage{times}
\usepackage{latexsym}
\usepackage{emnlp2016}

\newcommand{\bmx}[0]{\begin{bmatrix}}
\newcommand{\emx}[0]{\end{bmatrix}}

\newcommand{\vect}[1]{\mathbf{#1}}
\newcommand{\vects}[1]{\boldsymbol{#1}}

\newcommand{\vc}[0]{\vect{c}}

\newcommand{\vh}[0]{\vect{h}}

\newcommand{\vz}[0]{\vect{z}}

\newcommand{\TT}[0]{\vects{\theta}}

\newcommand{\HH}[0]{\mathcal{H}}

\newcommand{\enc}[0]{\ensuremath{\text{enc}}}
\newcommand{\att}[0]{\ensuremath{\text{att}}}
\newcommand{\dec}[0]{\ensuremath{\text{dec}}}
\newcommand{\oout}[0]{\ensuremath{\text{oout}}}
\newcommand{\iinit}[0]{\ensuremath{\text{init}}}

\DeclareMathOperator*{\argmax}{\arg \max}

\emnlpfinalcopy

\def\emnlppaperid{***}

\title{Can neural machine translation do simultaneous translation?}

\author{Kyunghyun Cho \and Masha Esipova \\
    New York University \\
{\tt \{kyunghyun.cho,masha.esipova\}@nyu.edu}}

\date{}

\begin{document}

\maketitle

\begin{abstract}
    We investigate the potential of attention-based neural machine translation
    in simultaneous translation. We introduce a novel decoding algorithm, called
    {\it simultaneous greedy decoding}, that allows an existing neural machine
    translation model to begin translating before a full source sentence is
    received. This approach is unique from previous works on
    simultaneous translation in that segmentation and translation are done
    jointly to maximize the translation quality and that translating each
    segment is strongly conditioned on all the previous segments.  This paper
    presents
    a first step toward building a full simultaneous translation system based on
    neural machine translation.
\end{abstract}

\section{Introduction}

Simultaneous translation differs from a more usual consecutive translation. In
simultaneous translation, the objective of a translator, or a translation
system, is defined as a combination of quality and delay, as opposed
to consecutive translation in which translation quality alone matters. In order
to minimize delay while maximizing quality, a simultaneous
translator must start generating symbols in a target languages {\it before} a
full source sentence is received.

Conventional approaches to simultaneous translation divide the translation
process into two
stages~\cite{bangalore2012real,fujita2013simple,sridhar2013segmentation,yarmohammadi2013incremental}.
A segmentation algorithm, or model, first divides a source sentence into
phrases. Each phrase is then translated by an underlying, often black box,
translation system, largely independent of the preceding phrases. These two
stages are separate from each other, in that the exchange of information between
these two modules is limited.

In this paper, we study the problem of simultaneous translation in the context
of neural machine
translation~\cite{Forcada1997,sutskever2014sequence,bahdanau2014neural}. Rather
than attempting to build a completely novel model along with a new training
algorithm, we design a novel decoding algorithm, called {\it simultaneous greedy
decoding}, that is capable of performing simultaneous translation with a neural
machine translation model trained to maximize the quality of consecutive
translation. Unlike previous approaches, our proposal performs segmentation and
translation jointly based solely on the translation quality (indirectly measured
by the conditional probabilities.) Furthermore, translation of each and every
segment is fully conditioned on all the preceding segments through the hidden
states of a recurrent network.

We extensively evaluate the proposed simultaneous greedy decoding together with
two waiting criteria on three language pairs--En-Cs, En-De and En-Ru. Our
analysis reveals that it is indeed possible to use an existing neural machine
translation system for simultaneous translation, and the proposed algorithm
provides a way to control the trade-off between quality and delay. Our
qualitative analysis on En-Ru simultaneous translation reveals interesting
behaviours such as phrase repetition as well as premature commitment.

We consider this work as a first step toward building a full simultaneous
translation system based on neural machine translation. In the conclusion, we
identify the following directions for the future research. First, a trainable
waiting criterion will allow a deeper integration between simultaneous decoding
and neural machine translation, resulting in a better simultaneous translation
system. Second, we need to eventually develop a learning algorithm specifically
for simultaneous translation.

\section{Attention-based Neural Translation}

Neural machine
translation~\cite{Forcada1997,kalchbrenner2013recurrent,sutskever2014sequence,bahdanau2014neural},
has recently become a major alternative to the existing statistical phrase-based
machine translation system~\cite{koehn2003statistical}. For instance, in the
translation task of WMT'16, the top rankers for En$\leftrightarrow$Cs,
En$\leftrightarrow$De, En$\to$Fi and En$\to$Ru all used attention-based neural
machine translation~\cite{bahdanau2014neural,luong2015effective}.\footnote{
    newstest2016 at \url{http://matrix.statmt.org/}
}

The attention-based neural machine translation is built as a composite of three
modules--encoder, decoder and attention mechanism. The encoder is usually
implemented as a recurrent network that reads a source sentence $X=(x_1, \ldots,
x_{T_x})$ and returns a set of context vectors $C=\left\{ \vh_1, \ldots,
\vh_{T_X} \right\}$, where
\begin{align}
    \label{eq:unienc}
    \vh_t = \phi_{\enc}(\vh_{t-1}, x_t).
\end{align}
Instead of a vanilla, unidirectional recurrent
network~\cite{luong2015effective}, it is possible to use a more sophisticated
network such as a bidirectional recurrent network~\cite{bahdanau2014neural} or a
tree-based recursive network~\cite{eriguchi2016tree}.

The decoder is a conditional language model based on a recurrent
network~\cite{mikolov2010recurrent}. At each time step $t'$, it first uses the
attention mechanism to compute a single time-dependent vector out of all
the context vectors:
$\vc_{t'} = \sum_{t=1}^{T_x} \alpha_t \vh_t$,
where 
\begin{align}
    \label{eq:alignment}
    \alpha_t \propto \exp\left( f_{\att}(\vz_{t'-1}, \tilde{y}_{t'-1}, \vh_t)
    \right).
\end{align}
$\vz_{t-1}$ and $\tilde{y}_{t-1}$ are the decoder's hidden state and the
previous target word. This content-based attention mechanism can be
extended to incorporate also the location of each context
vector~\cite{luong2015effective}.

The decoder updates its own state with this time-dependent context vector by
$
    \vz_{t'} = \phi_{\dec}(\vz_{t'-1}, \tilde{y}_{t'-1}, \vc_{t'}).
$
The distribution over the next target word is then computed by
\begin{align*}
    p(y_{t'} = j|\tilde{y}_{<t'}, X) \propto \exp\left( 
        \phi_{\oout}(\vz_{t'})
    \right).
\end{align*}
The initial state of the decoder is often initialized as
\begin{align}
    \label{eq:dec_init}
    \vz_0 = \phi_{\iinit}(C).
\end{align}

The whole model, consisting of the encoder, decoder and attention mechanism, is
jointly trained to maximize the log-likelihood given a set of $N$ training pairs:
$
    \max_{\TT} \frac{1}{N} \sum_{n=1}^N \sum_{t'=1}^{T_{y}^n} \log
    p(y_{t'}^n|y_{<t'}^n, X^n),
    $
where $\TT$ denotes all the parameters.

\begin{algorithm}[t]
    \caption{Simultaneous Greedy Decoding}
    \label{alg:simul_decode}
    \begin{algorithmic}[1]
        \REQUIRE $\delta$, $s_0$, Input Pipe $X$, Output Pipe $Y$
        \STATE Initialize $s \leftarrow s_0$, $C \leftarrow \text{READ}(X, s)$, $C' \leftarrow \left\{ \right\}$
        \STATE Initialize the decoder's state $\vz_0$ based on $C$
        \WHILE{\TRUE}
            \STATE $\hat{y}_t = \argmax_{y_t} \log p(y_t | y_{<t}, C)$
            \IF{$s\geq T_{X}$} 
                \STATE $\text{WRITE}(Y, \hat{y}_t)$, $t \leftarrow t + 1$
            \ELSE
                \STATE $C' \leftarrow \text{READ}(X, \delta)$ {\bf if} $|C'| = 0$.
                \IF{$\Lambda(C, C \cup C')$}
                    \STATE $C \leftarrow C \cup C'$, $s \leftarrow s + \delta$,
                    $C' \leftarrow \left\{ \right\}$
                    \STATE {\bf continue}
                \ELSE
                    \STATE $\text{WRITE}(Y, \hat{y}_t)$, $t \leftarrow t + 1$
                \ENDIF
            \ENDIF
            \IF{$\hat{y}_t = \left<\text{eos}\right>$}
            \STATE {\bf break}
            \ENDIF
        \ENDWHILE
    \end{algorithmic}
\end{algorithm}

\section{Simultaneous Greedy Decoding}

We investigate the potential of using a {\it trained} neural machine translation
as a simultaneous translator by introducing a novel decoding algorithm. In this
section, we propose and describe such an algorithm, called {\it simultaneous greedy
decoding}, in detail.

\subsection{Algorithm}

An overall algorithm is presented in Alg.~\ref{alg:simul_decode}. 

\paragraph{Input Arguments}

The proposed simultaneous greedy decoding has two hyperparameters that are used
to control the trade-off between delay and quality. They are the step size
$\delta$ and the number of initially-read source symbols $s_0$. We will describe
how the delay is defined later in Sec.~\ref{sec:delay}.

As the goal is to do simultaneous translation, the algorithm receives two input
$X$ and output $Y$ pipes instead of a full source sentence at the beginning.
Each pipe comes with two API's which are $\text{WRITE}$ and $\text{READ}$.
$\text{WRITE}(P, x)$ commits the symbol $x$ to the pipe $P$, and $\text{READ}(P,
n)$ receives $n$ symbols from the pipe $P$ and computes the additional
context vectors using the encoder (line 11 of
Alg.~\ref{alg:simul_decode}).

\paragraph{State}

The proposed algorithm maintains the following state variables:
\vspace{-3mm}
\begin{enumerate}
    \itemsep -.5em
    \item $s$: \# of received source words
    \item $t$: \# of committed target words + 1
    \item $C=\left\{ \vh_1, \ldots, \vh_s \right\}$
    \item $C'$: additional context vectors
    \item $\vz_{t-1}$: the decoder's latest hidden state
\end{enumerate}

\paragraph{Initialization}

Initially, the algorithm reads $s_0$ source symbols from $X$ and initialize its
context set $C$ to contain the context vectors $\vh_1,\ldots,\vh_{s_0}$. Based
on this initial context set, the decoder's hidden state $\vz_0$ is initialized
according to Eq.~\eqref{eq:dec_init}. See lines 2--3. 

\paragraph{Translation Core}

At each iteration, the algorithm checks whether the full source sentence has
been read already (line 6). If so, it simply performs greedy decoding by
committing the most likely target symbol to $Y$, as would have been done in
consecutive translation, while incrementing $t$ by one for each committed
target word (line 7).  

Otherwise, the algorithm reads $\delta$ more source symbols from $X$ and forms
an additional context set $C'$, if $C'$ is currently empty. It then compares the
conditional distributions over the next target symbols $y_t$ given either the
current context set $C$ or the new context set $C \cup C'$ based on a predefined
criterion $\Lambda$. The criterion should be designed so as to determine whether
the amount of information in $C$ is enough to make a decision in that the
additional information $C'$ would not make a difference. Two such criteria will
be described later in Sec.~\ref{sec:criteria}

If this waiting criterion is met (i.e., it is determined better to consider
$C'$,) the context set is updated to include the additional context vectors
(line 13,) and $C'$ is flushed. If not, the algorithm commits the most likely
target symbols $\hat{y}_t$ given the current context set $C$ and increments $t$
by one (line 15). 

The iteration terminates when the last committed symbol $\hat{y}_t$ is the
end-of-sentence symbol. 

\paragraph{Computational Complexity}

The worse-case complexity of this simultaneous greedy decoding is twice as
expensive as that of a consecutive greedy decoding is, 
    with some trivial cacheing of $C'$, $\log p(y_t|y_{<t}, C)$ and
    $\log p(y_t|y_{<t}, C \cup C')$.
The worse case happens when the waiting criterion is met for all the source
symbols, in which case the proposed algorithm reduces to the consecutive
greedy translation. The complexity however drops significantly in practice as
the waiting criterion is satisfied more, because this reduces the number of
context vectors over which the alignment weights are computed (see
Eq.~\eqref{eq:alignment}.) The empirical evaluation, which will be presented
later in this paper, reveals that we can control this behaviour by adjusting
$\delta$ and $s_0$.

\paragraph{Why ``Greedy'' Decoding?}

It is possible to extend the proposed algorithm to be less greedy. For
instance, instead of considering a single target word at a time ($\hat{y}_t$),
we can let it consider a multi-symbol sequence at a time. This however increase
the computational complexity at each time step greatly, and we leave it as a
future research.

\subsection{Waiting Criteria $\Lambda$}
\label{sec:criteria}

At each point in time, the simultaneous greedy decoding makes a decision on
whether to wait for a next batch of source symbols or generate a target symbol
given the current context ($C \cup C'$ in line 10 of
Alg.~\ref{alg:simul_decode}.) This decision is made based on a predefined
criterion $\Lambda(C, C\cup C')$.
In this paper, we investigate two different, but related criteria. 

\paragraph{Criterion 1: Wait-If-Worse}

The first alternative is {\bf Wait-If-Worse}. It considers whether the
confidence in the prediction based on the current source context (up to $s$
source words) decreases with more source context (up to $s+\delta$ source
words). This is detected by comparing the log-probabilities of the target word
selected to be most likely when only the first $s$ source words were considered,
i.e.,
\begin{align}
    \label{eq:wait-if-worse}
    \Lambda(C, C\cup C'):
    (&\log  p(\hat{y}|\hat{y}_{<t}, C) \\
                    &> \log
    p(\hat{y}|\hat{y}_{<t},C\cup C')),
    \nonumber
\end{align}
where $\hat{y} = \argmax_{y} p(y|\hat{y}_{<t}, C)$.

\paragraph{Criterion 2: Wait-If-Diff}

The next alternative, {\bf Wait-If-Diff}, compares the most likely target
words given up to $s$ and up to $s+\delta$ source words. If they are same, we
commit to this target word. Otherwise, we wait. This is different from the
Wait-If-Worse criterion, because the decrease in the log-probability of the most
likely word given more source words does not necessarily imply that the most
likely word has changed. We define this criterion as 
\begin{align}
    \label{eq:wait-if-diff}
    \Lambda(C, C \cup C'): (\hat{y} \neq \hat{y}'),
\end{align}
where 
$
    \hat{y}' = \argmax_{y} \log p(y| \hat{y}_{<t}, C \cup C').
$

\begin{figure*}[t]
    \small

    \begin{minipage}{0.33\textwidth}
        \centering
        Czech (12.12m)
    \end{minipage}
    \begin{minipage}{0.33\textwidth}
        \centering
        German (4.15m)
    \end{minipage}
    \begin{minipage}{0.33\textwidth}
        \centering
        Russian (2.32m)
    \end{minipage}
    \vspace{-7mm}

    \begin{minipage}{0.33\textwidth}
        \centering
        \includegraphics[width=1\linewidth]{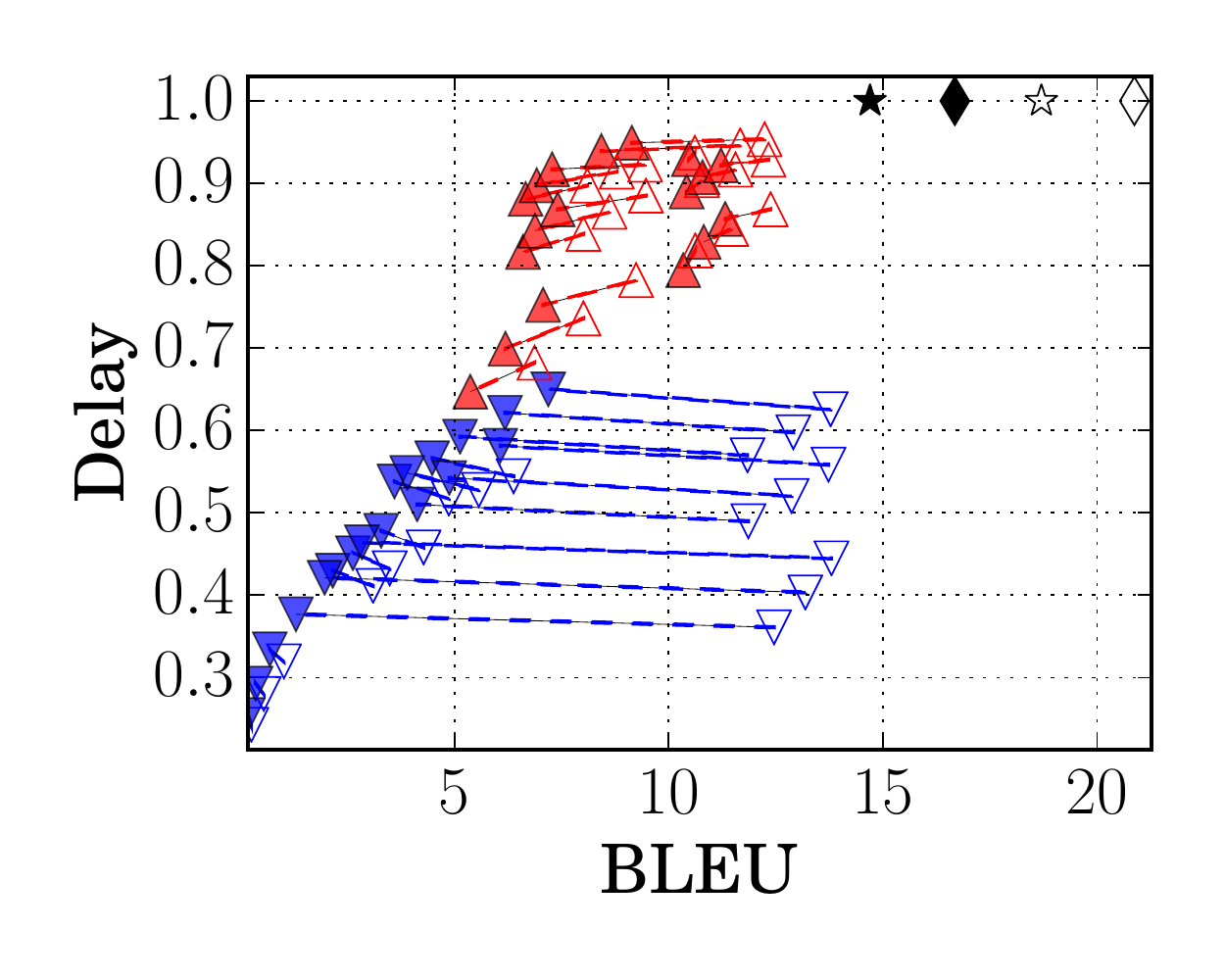}
    \end{minipage}
    \begin{minipage}{0.33\textwidth}
        \centering
        \includegraphics[width=1\linewidth]{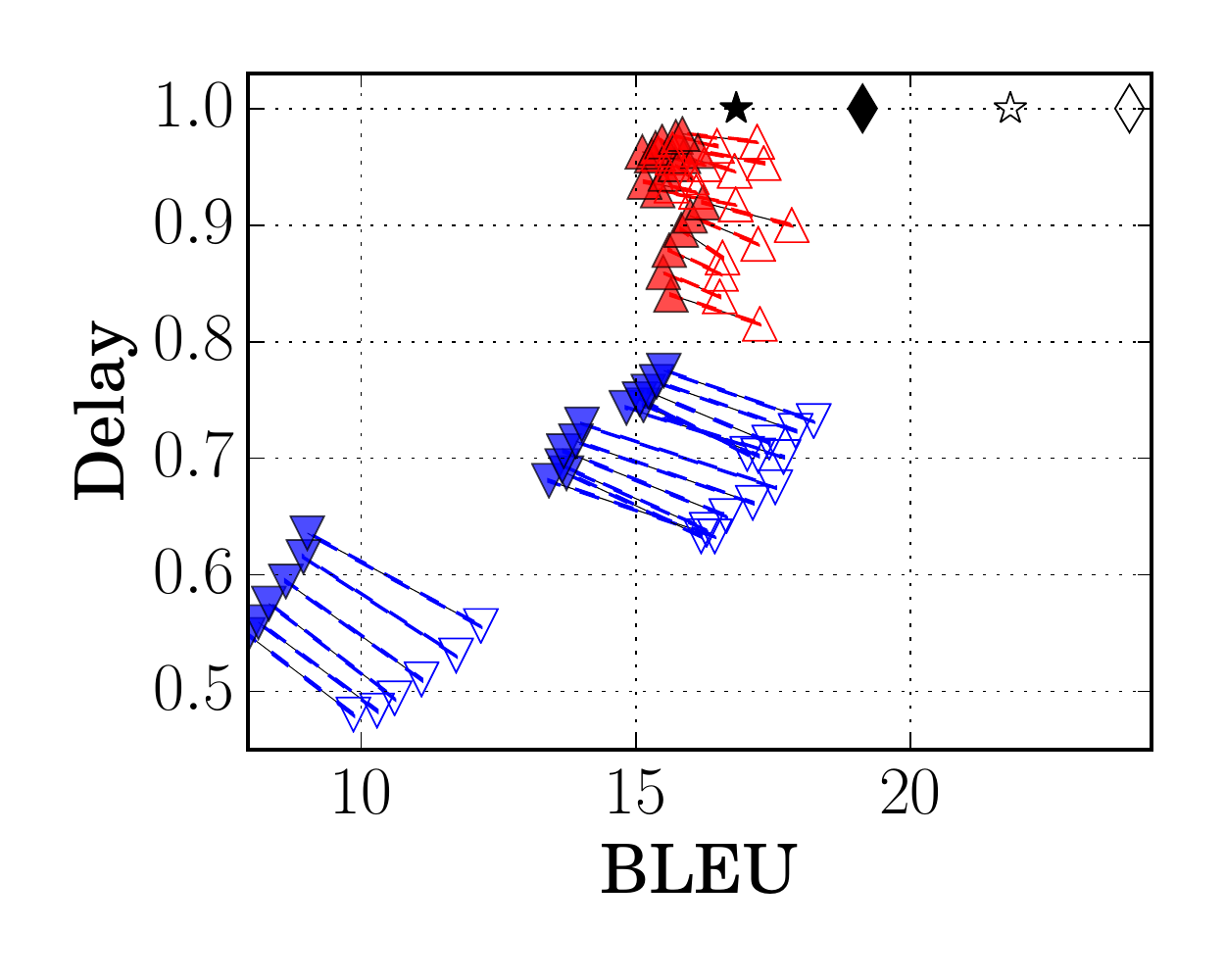}
    \end{minipage}
    \begin{minipage}{0.33\textwidth}
        \centering
        \includegraphics[width=1\linewidth]{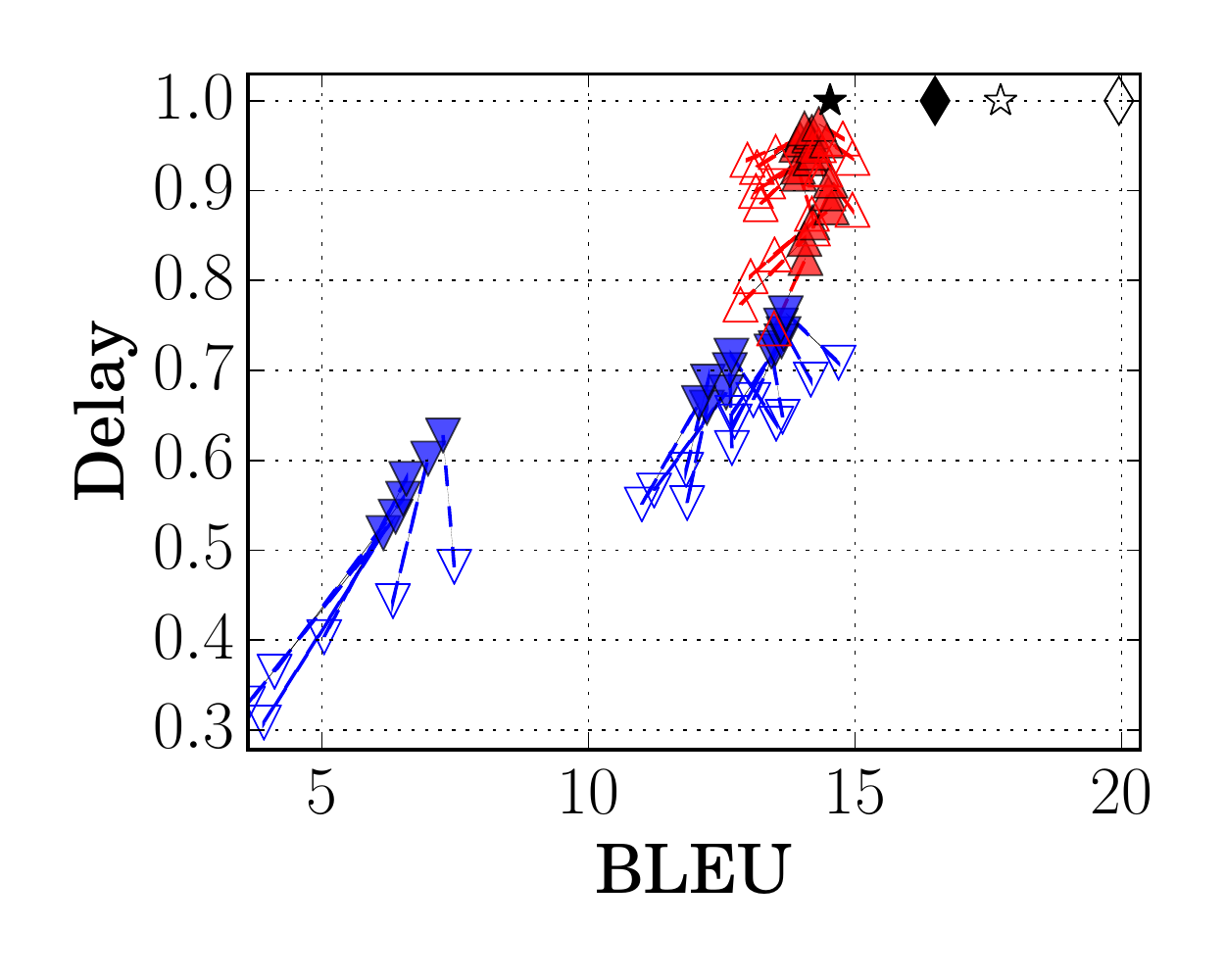}
    \end{minipage}

    \vspace{-3mm}
    \caption{Quality vs. Delay $\tau$ plots for all the language
        pair--directions. 
        ${\color{red} \blacktriangle}$: Wait-If-Worse (En$\to$). 
        ${\color{blue} \blacktriangledown}$: Wait-If-Diff (En$\to$).
        ${\color{red} \triangle}$: Wait-If-Worse ($\to$En). 
        ${\color{blue} \triangledown}$: Wait-If-Diff ($\to$En).
        $\bigstar$: consecutive greedy decoding (En$\to$). 
        $\blacklozenge$: consecutive beam search (En$\to$).
        \ding{73}: consecutive greedy decoding ($\to$En). 
        $\lozenge$: consecutive beam search ($\to$En).
        Each dashed line connects the points with the same decoding
        parameters ($\delta$ and $s_0$) between translating to and from English.
        Delay $\tau$: Lower the better.
        BLEU: Higher the better.
    }
    \label{fig:overall}

    \vspace{-4mm}
\end{figure*}

\paragraph{Other Criteria}

Although we consider the two criteria described above, it is certainly possible
to design another criterion. For instance, we have tested the following
entropy-based criterion during our preliminary experiments:
\[
    \Lambda(C, C \cup C'): (\HH(y|\hat{y}_{<t}, C) > \HH(y|\hat{y}_{<t}, C\cup
    C')).
\]
That is, the algorithm should wait for next source symbols, if the entropy is
expected to decrease (i.e., if the confidence in prediction is expected to
increase.) This criterion however did not work as well. 

\subsection{Delay in Translation}
\label{sec:delay}

For each decoded target symbol $\hat{y}_t$, we can track how many source symbols
were required, i.e., $|C \cup C'|$ in line 10 of Alg.~\ref{alg:simul_decode}. We
will use $s(t)$ to denote this quantity. Then, we define the (normalized) total
amount of time spent for translating a give source sentence, or equivalent {\it
delay in translation}, as
\begin{align}
    \label{eq:delay}
    0 < \tau(X, \hat{Y}) = \frac{1}{|X||\hat{Y}|} \sum_{t=1}^{|\hat{Y}|} s(t)
    \leq 1.
\end{align}

In the case of full translation, as opposed to simultaneous translation,
$\tau(X, Y)=1$ for all $Y$. This follows from the fact that $s(t)=|X|$ for all
$t$. In the case of word-by-word translation in which case $|X|=|Y|$, $\tau(X,
Y)=0.5$, because $s(t)=t$. 


\paragraph{Alignment vs. Delay}

$s(t)$ however does not accurately reflect on which source prefix the $t$-th
target symbol is conditioned.
It is rather $s'(t)=|C|$ instead, which reflects the alignment between the
source and target symbols. It is thus more useful to
inspect $s'(t)$ in order to understand the behaviour of the proposed
simultaneous greedy decoding.  

\begin{figure*}[t]
    \small

    \begin{minipage}{0.01\textwidth}
        ~
    \end{minipage}
    \begin{minipage}{0.49\textwidth}
        \centering
        En$\to$
    \end{minipage}
    \begin{minipage}{0.49\textwidth}
        \centering
        $\to$En
    \end{minipage}
    \vspace{-6mm}

    \begin{minipage}{0.01\textwidth}
        \rotatebox[origin=c]{90}{Czech}
    \end{minipage}
    \begin{minipage}{0.49\textwidth}
        \centering
        \includegraphics[width=\linewidth]{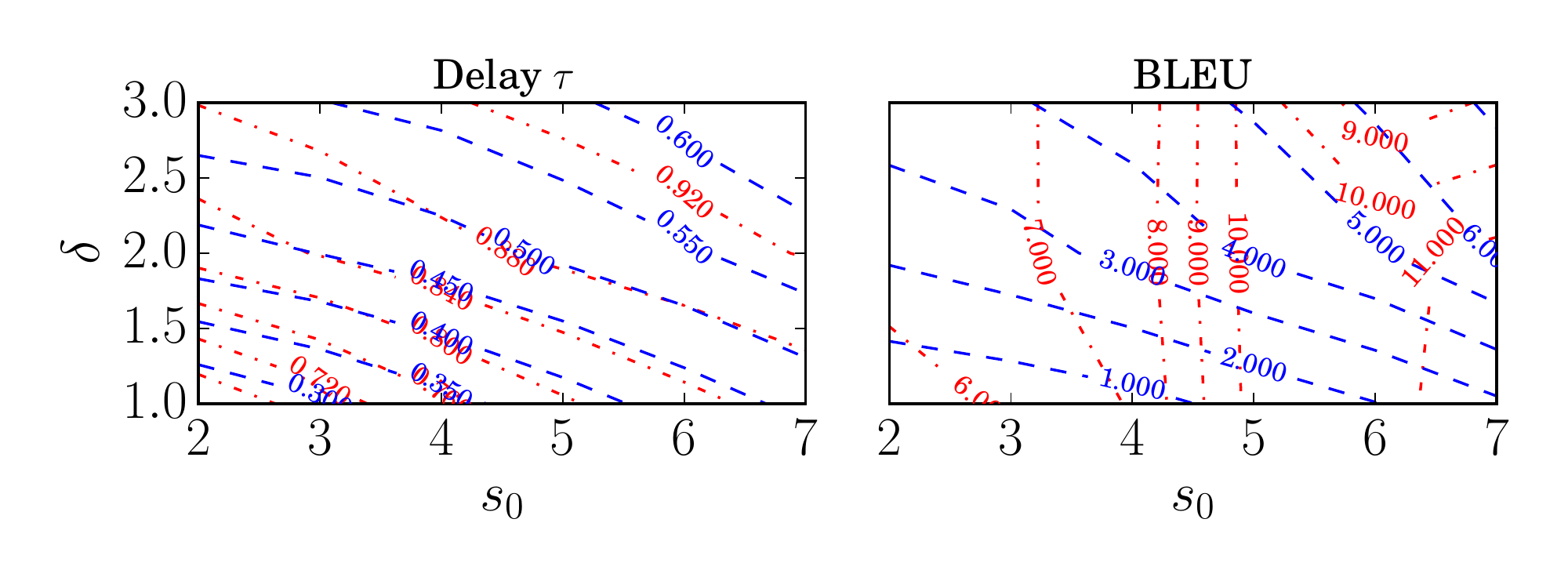}
    \end{minipage}
    \begin{minipage}{0.49\textwidth}
        \centering
        \includegraphics[width=\linewidth]{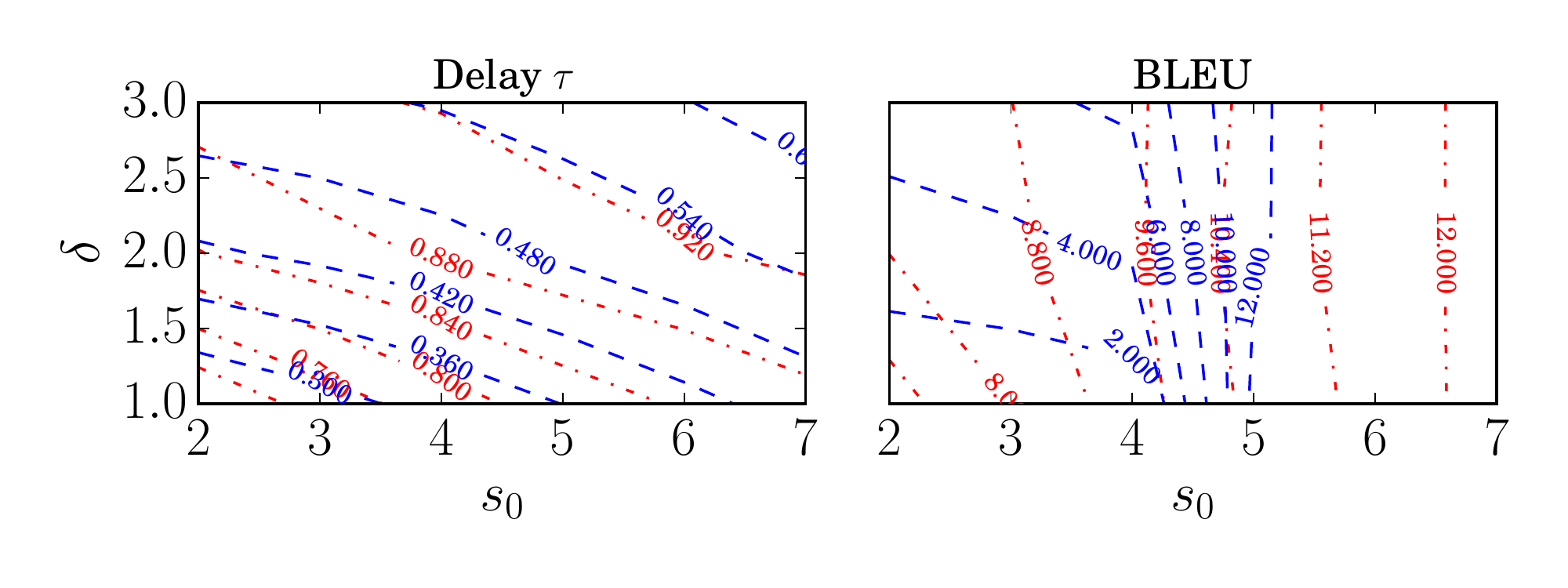}
    \end{minipage}
    \vspace{-5mm}

    \begin{minipage}{0.01\textwidth}
        \rotatebox[origin=c]{90}{German}
    \end{minipage}
    \begin{minipage}{0.49\textwidth}
        \centering
        \includegraphics[width=\linewidth]{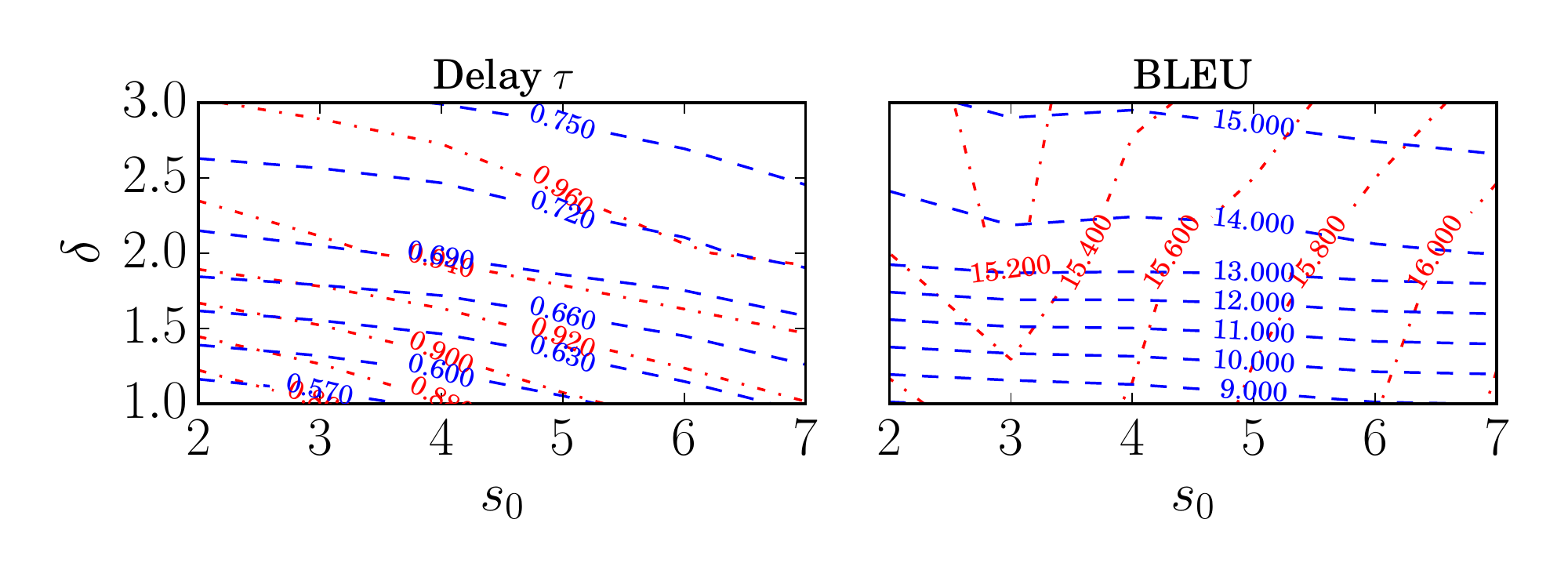}
    \end{minipage}
    \begin{minipage}{0.49\textwidth}
        \centering
        \includegraphics[width=\linewidth]{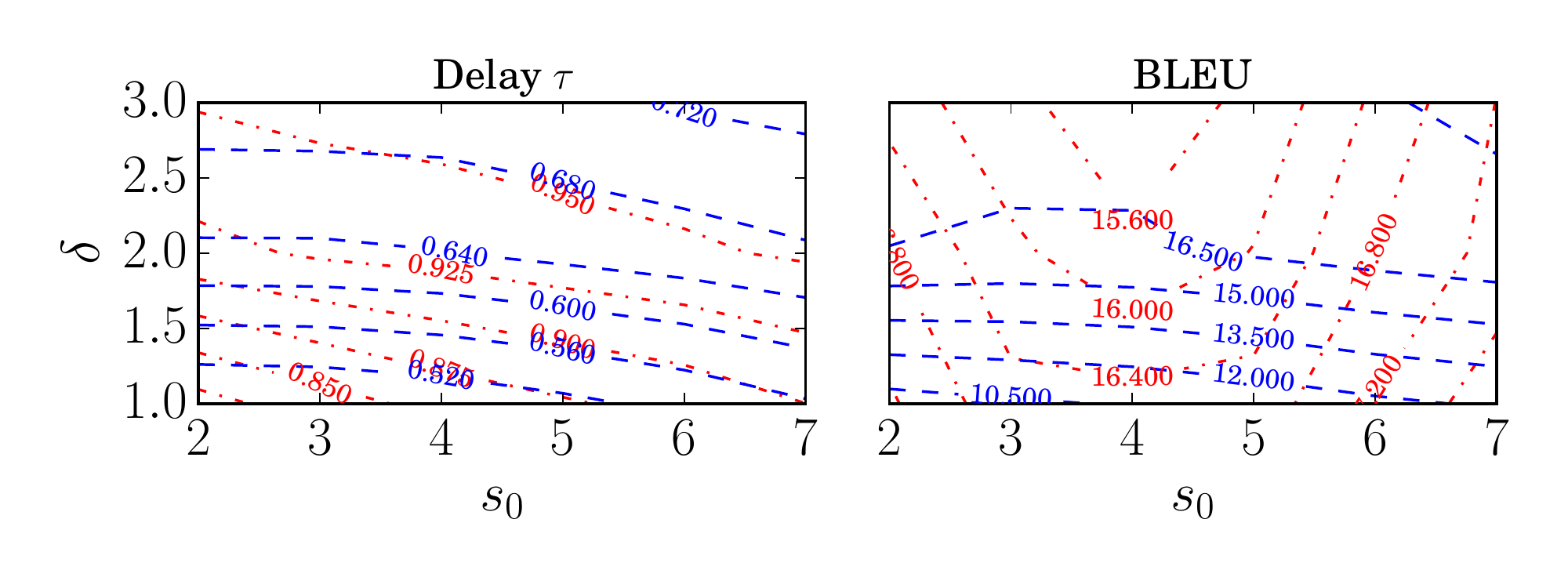}
    \end{minipage}
    \vspace{-5mm}

    \begin{minipage}{0.01\textwidth}
        \rotatebox[origin=c]{90}{Russian}
    \end{minipage}
    \begin{minipage}{0.49\textwidth}
        \centering
        \includegraphics[width=\linewidth]{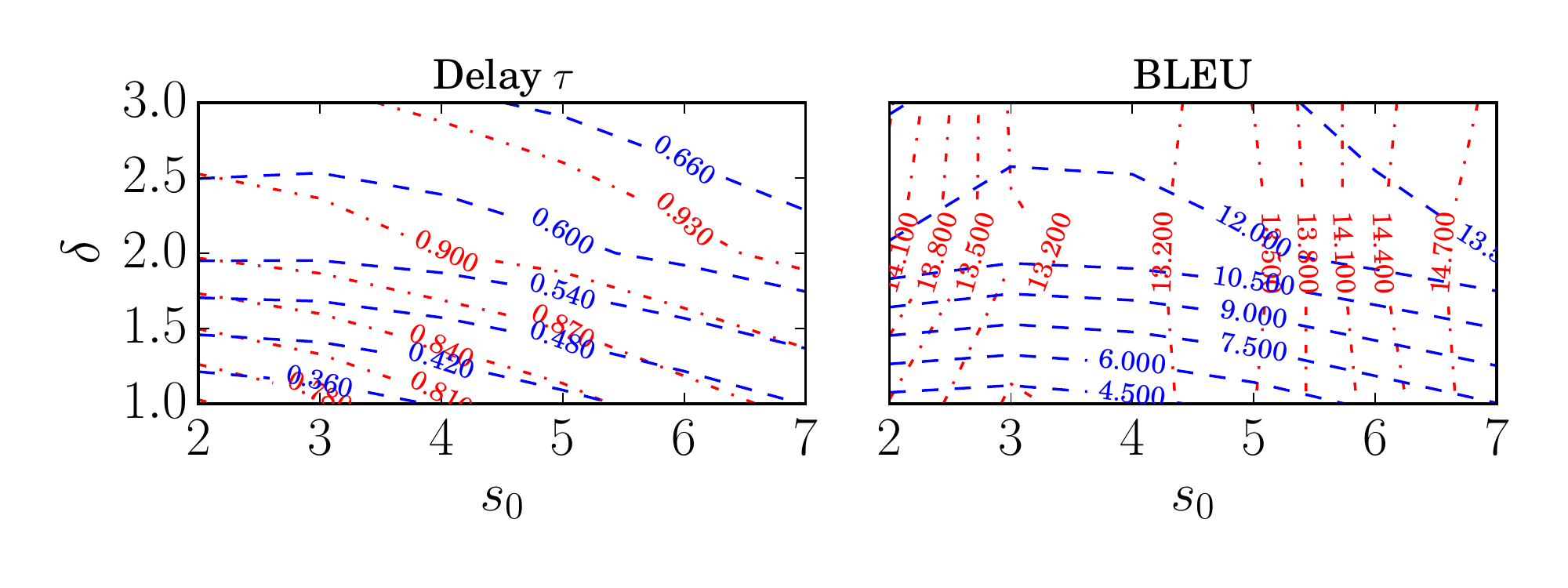}
    \end{minipage}
    \begin{minipage}{0.49\textwidth}
        \centering
        \includegraphics[width=\linewidth]{both_contour_ru_to-en.pdf}
    \end{minipage}

    \vspace{-3mm}
    \caption{Quality and delay $\tau$ per $s_0$ and
        $\delta$. Red dash-dot curves ({\scriptsize ${\color{red} -\cdot -}$}): Wait-If-Worse. 
        Blue dashed curves ({\scriptsize ${\color{blue} --}$}): Wait-If-Diff.
    }
    \label{fig:criteria}

    \vspace{-4mm}
\end{figure*}

\section{Related Work}

Perhaps not surprisingly, there are only a small number of prior work in {\it
simultaneous} machine translation. Much of those works are done in the context
of speech
translation~\cite{bangalore2012real,fujita2013simple,sridhar2013segmentation,yarmohammadi2013incremental}.
Often, incoming speech is transcribed by an automatic speech recognition (ASR)
system, and the transcribed text is segmented into a translation unit largely
based on acoustic and linguistic cues (e.g., silence or punctuation marks.) Each
of these segments is then translated largely independent from each other by a
separate machine translation system. The simultaneous greedy decoding with
neural machine translation, proposed in this paper, is clearly distinguished
from these approaches in that (1) segmentation and translation happen jointly to
maximize the translation quality and (2) each segmentation is strongly dependent
on all the previous segment in both the source and target sentences.

More recently, \newcite{grissom2014don} and \newcite{oda2014optimizing} proposed
to extend those earlier approaches by introducing a trainable segmentation
policy which is trained to maximizes the translation quality. For instance, the
trainable policy introduced in \cite{grissom2014don} works by keeping an
intermediate source/translation prefix, querying an underlying black-box machine
translation system and deciding to commit these intermediate translations once a
while. The policy is trained in the framework of imitation
learning~\cite{daume2009search}. In both of these cases translation is still
done largely per segment, unlike the simultaneous translation with neural
machine translation. We however find this idea of  learning a policy for
simultaneous decoding be directly applicable to the proposed simultaneous greedy
decoding and leave it as a future work.

Another major contribution by \newcite{grissom2014don} was to let the policy
{\it predict} the final verb of a source sentence, which is especially useful in
the case of translating from a verb-final language (e.g., German) to another
type of language. They do so by having a separate verb-conditioned $n$-gram
language model of all possible source prefixes. This prediction, explicit in
\cite{grissom2014don} is however done implicitly in neural machine translation,
where the decoder acts as strong language model. 

\begin{figure*}[t]
    \small

    \begin{minipage}{0.01\textwidth}
        ~
    \end{minipage}
    \begin{minipage}{0.49\textwidth}
        \centering
        Wait-If-Worse
    \end{minipage}
    \begin{minipage}{0.49\textwidth}
        \centering
        Wait-If-Diff
    \end{minipage}
    \vspace{-5mm}

    \begin{minipage}{0.01\textwidth}
        \rotatebox[origin=c]{90}{$\to$Cz}
    \end{minipage}
    \begin{minipage}{0.49\textwidth}
        \centering
        \includegraphics[width=\linewidth,clip=True,trim=0 0 0 0]{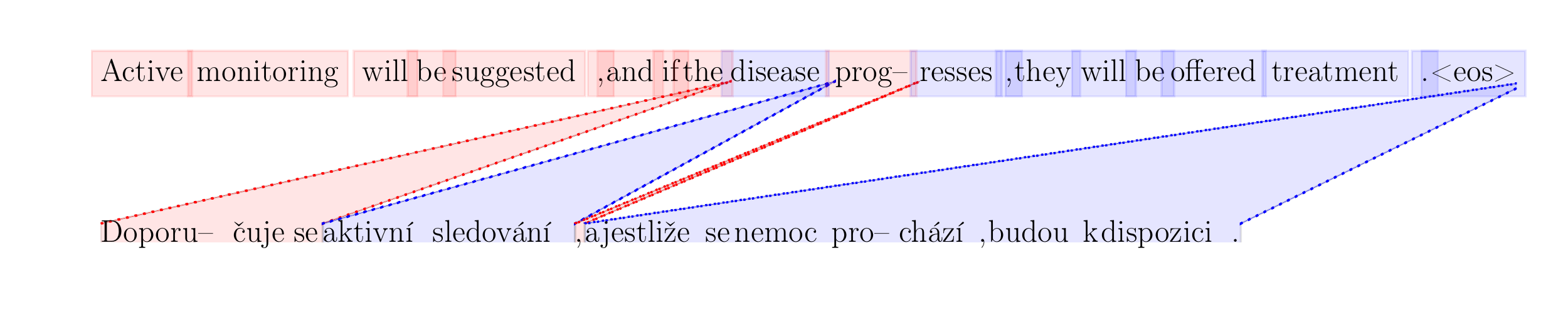}
    \end{minipage}
    \begin{minipage}{0.49\textwidth}
        \centering
        \includegraphics[width=\linewidth,clip=True,trim=0 0 0 0]{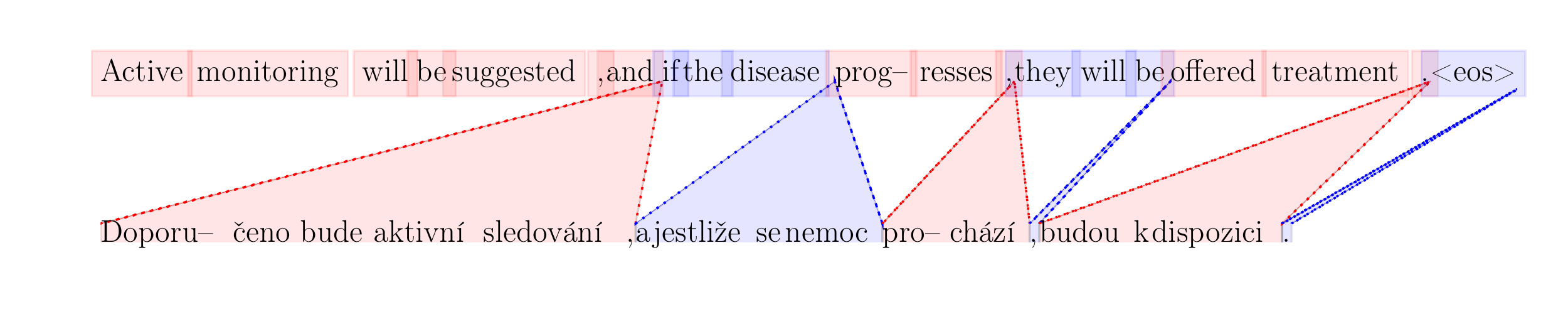}
    \end{minipage}
    \vspace{-5mm}

    \begin{minipage}{0.01\textwidth}
        \rotatebox[origin=c]{90}{$\to$De}
    \end{minipage}
    \begin{minipage}{0.49\textwidth}
        \centering
        \includegraphics[width=\linewidth,clip=True,trim=0 0 0 0]{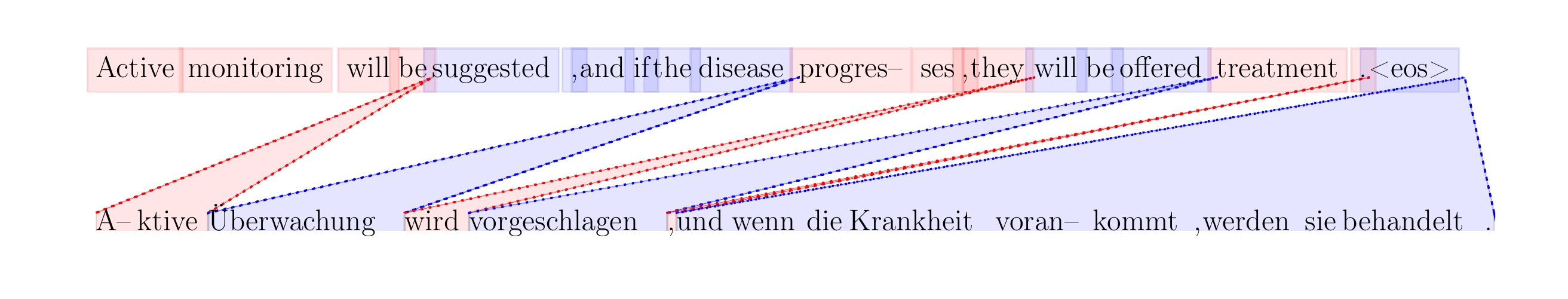}
    \end{minipage}
    \begin{minipage}{0.49\textwidth}
        \centering
        \includegraphics[width=\linewidth,clip=True,trim=0 0 0 0]{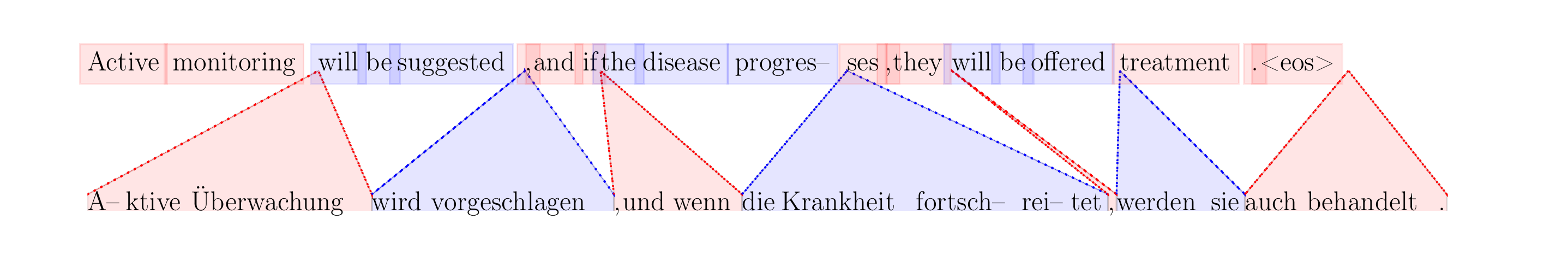}
    \end{minipage}
    \vspace{-5mm}


    \begin{minipage}{0.01\textwidth}
        \rotatebox[origin=c]{90}{~}
    \end{minipage}
    \begin{minipage}{0.98\textwidth}
        \centering
        (a) From English: ``{\it\scriptsize Active monitoring will
        be suggested , and if the disease progresses , they will be offered
    treatment .}''
    \end{minipage}

    \begin{minipage}{0.01\textwidth}
        \rotatebox[origin=c]{90}{Cs$\to$}
    \end{minipage}
    \begin{minipage}{0.49\textwidth}
        \centering
        \includegraphics[width=\linewidth,clip=True,trim=0 0 0 0]{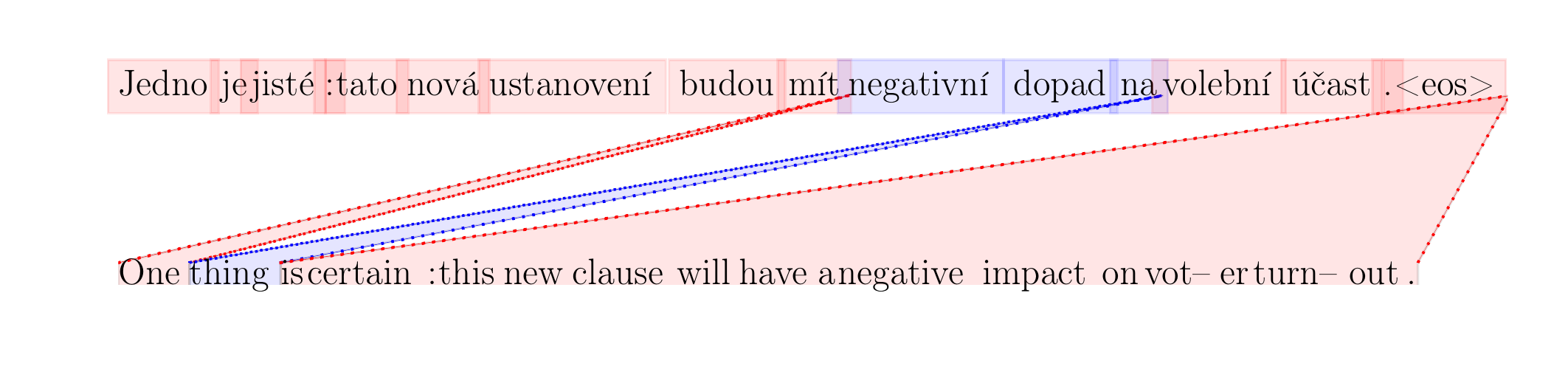}
    \end{minipage}
    \begin{minipage}{0.49\textwidth}
        \centering
        \includegraphics[width=\linewidth,clip=True,trim=0 0 0 0]{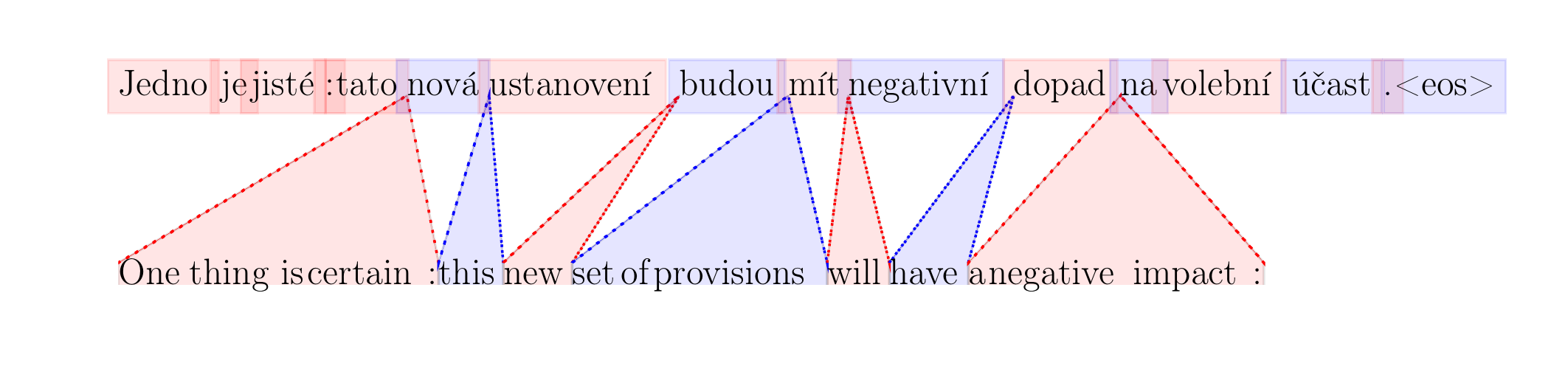}
    \end{minipage}
    \vspace{-5mm}

    \begin{minipage}{0.01\textwidth}
        \rotatebox[origin=c]{90}{De$\to$}
    \end{minipage}
    \begin{minipage}{0.49\textwidth}
        \centering
        \includegraphics[width=\linewidth,clip=True,trim=0 0 0 0]{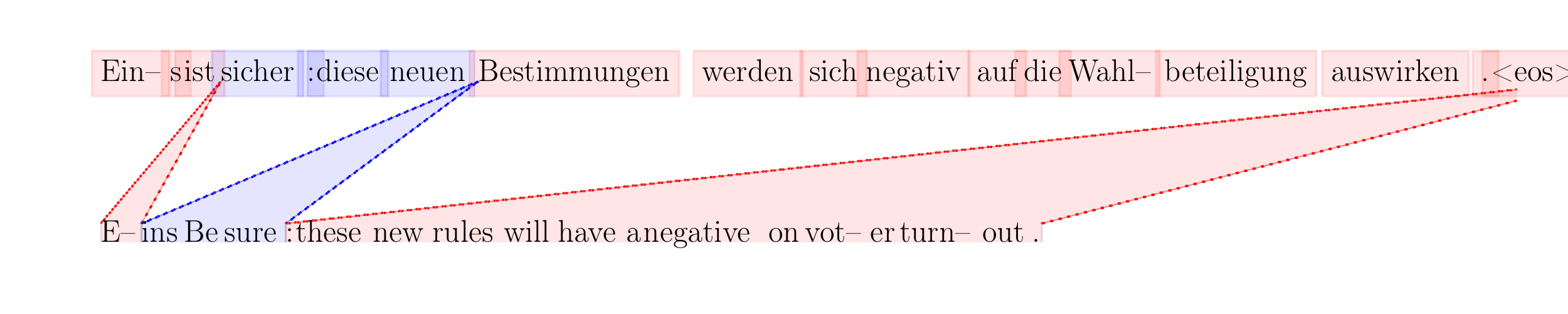}
    \end{minipage}
    \begin{minipage}{0.49\textwidth}
        \centering
        \includegraphics[width=\linewidth,clip=True,trim=0 0 0 0]{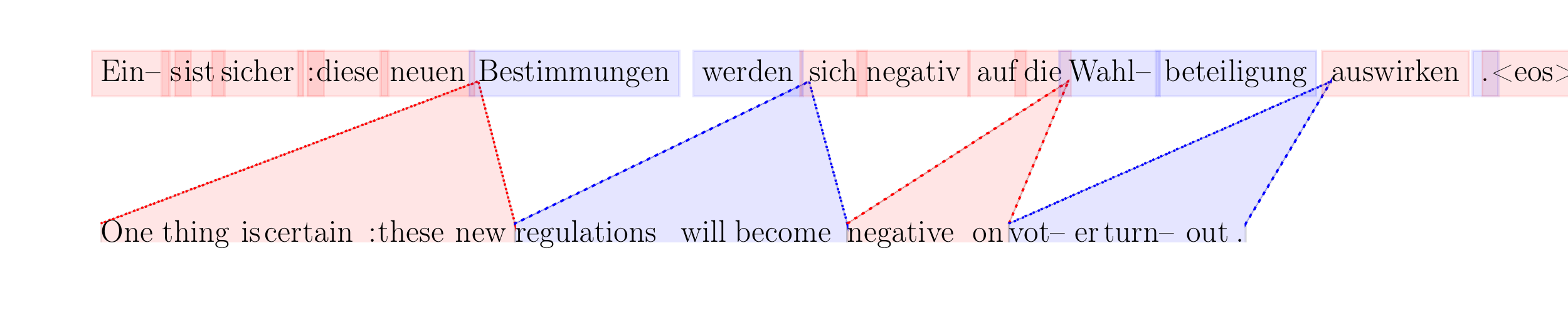}
    \end{minipage}
    \vspace{-5mm}


    \begin{minipage}{0.01\textwidth}
        \rotatebox[origin=c]{90}{~}
    \end{minipage}
    \begin{minipage}{0.98\textwidth}
        \centering
        (b) To English: ``{\it\scriptsize One thing is certain : these new provisions will
        have a negative impact on voter turn-out .}''
    \end{minipage}

    \caption{
        Example translation by simultaneous greedy decoding (a) from English and
        (b) to English. The source and reference sentences were selected
        randomly from the development set.  The correspondences between chunks
        of consecutive symbols in the source and translation are highlighted by
        background color. The dotted lines indicate the latest source symbol up
        to which the context was taken ($s'(t)$ from Sec.~\ref{sec:delay}). Best
        viewed when zoomed digitally.
    }
    \label{fig:example-from-en}

    \vspace{-4mm}
\end{figure*}

\section{Experimental Settings}

\paragraph{Tasks and Corpora} 
We extensively study the proposed simultaneous greedy decoding algorithm on
three language pairs--En-Cs (12.12m sentence pairs), En-De (4.15m) and En-Ru
(2.32m)-- and in both directions per pair.
We use all the parallel corpora available from WMT'15.\footnote{
    \url{http://www.statmt.org/wmt15/}
} All the sentences were first tokenized\footnote{
    tokenizer.perl from Moses~\cite{koehn2007moses}.
} and segmented into subword units using byte pair encoding (BPE) following
\cite{sennrich2015neural}. During training, we only use sentence pairs, where
both sides are less than or equal to 50 BPE subword symbols long. 

newstest-2013 is used as a validation set both to early-stop training and to
extensively evaluate the proposed simultaneous greedy decoding algorithm.
We use newstest-2015 as a test set to confirm that the neural translation models
used in the experiments are reasonably well trained.

\paragraph{Translation Models}
In total, we train six separate neural translation models, one for each
pair--direction. We use a {\it unidirectional} recurrent network with 1028 gated
recurrent units (GRU, \cite{cho2014learning}) as an encoder.
A decoder is similarly a recurrent neural net with 1028 GRUs. The
soft-alignment function is a feedforward network with one hidden layer
consisting of 1028 $\tanh$ units. Each model is trained with
Adadelta~\cite{zeiler2012adadelta} until the average log-probability on the
validation set does not improve, which takes about a week per model.

These trained models do not achieve the state-of-the-art translation qualities,
as they do not exploit the ensemble technique~\cite{sutskever2014sequence} nor
monolingual corpus~\cite{sennrich2015improving}, both of which have been found
to be crucial in improving neural machine translation. Under these constraints,
however, the trained models translates as well as those reported earlier by, for
instance, \newcite{firat2016multi}, as can be seen in Table~\ref{tab:perf}. 

\begin{table}[t]
    \centering
    \small

    \begin{tabular}{c | c || c c c}
        & & Cs & De & Ru \\
        \toprule
        \multirow{2}{*}{\rotatebox[origin=c]{90}{En$\to$}} & Ours & 15.2 & 19.5 & 17.77 \\
                                                           & $\star$ & 13.84 & 21.75 & 19.54  \\
        \midrule
        \multirow{2}{*}{\rotatebox[origin=c]{90}{$\to$En}} & Ours & 20.47 & 23.96 & 22.27 \\
                                                           & $\star$ & 20.32 & 24 & 22.44  
    \end{tabular}
    \caption{
        BLEU scores on the test set (newstest-2015) obtained by (Ours) the
        models used in this paper and ($\star$) from \protect\cite{firat2016multi}.
        Although our models use a unidirectional recurrent net as an encoder,
        the translation qualities are comparable.
    }
    \label{tab:perf}

    \vspace{-8mm}
\end{table}

\paragraph{Decoding Parameters}
With the proposed simultaneous greedy decoding, we vary $\delta \in \left\{ 1,
2, 3 \right\}$ and $s_0 \in \left\{ 2, 3, 4, 5, 6, 7 \right\}$. For each
combination, we report both BLEU and the delay measure $\tau$ from
Eq.~\eqref{eq:delay}. In order to put the translation quality of the
simultaneous translation in perspective, we report BLEU scores from
consecutive translation with both greedy and beamsearch decoding. The beam
width is set to $5$, as used in \cite{chung2016character}.

\begin{figure*}[t]
    \small

    \centering
    \begin{minipage}{0.32\linewidth}
        \centering
        \includegraphics[width=\linewidth,clip=True,trim=40 0 60 0]{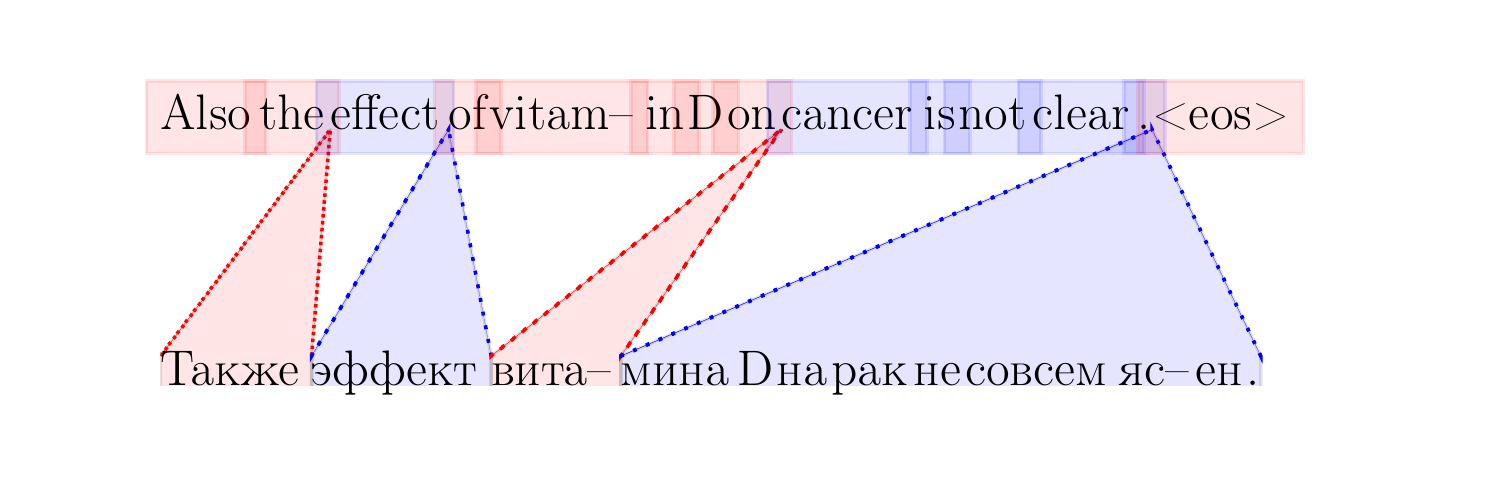}
    \end{minipage}
    \hfill
    \begin{minipage}{0.32\linewidth}
        \centering
        \includegraphics[width=\linewidth,clip=True,trim=40 0 60 0]{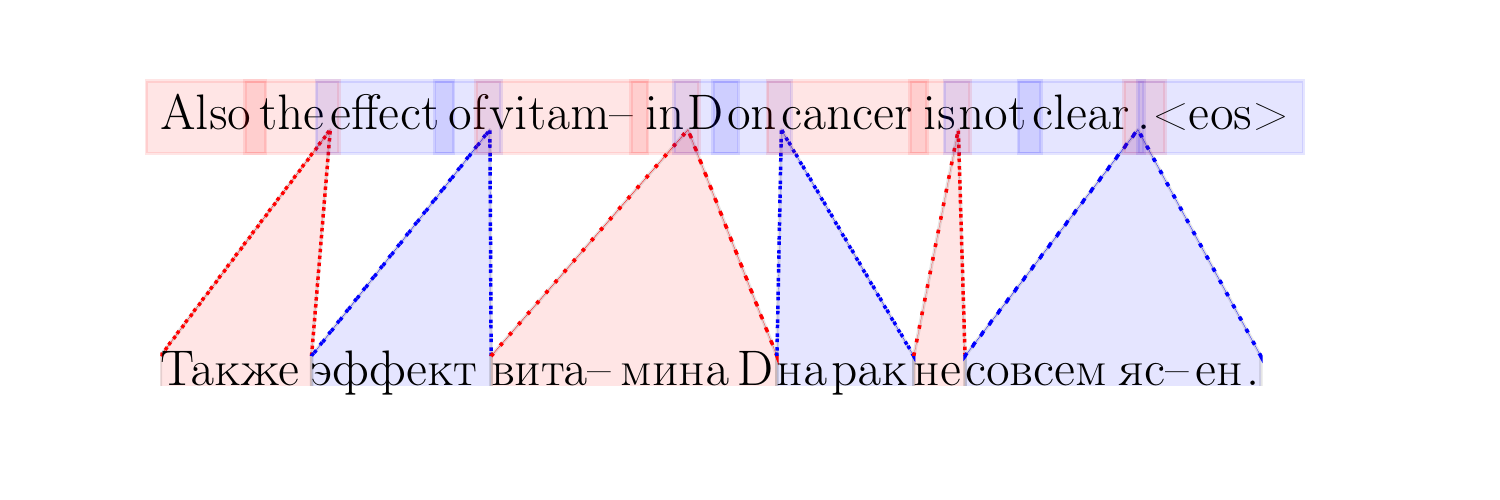}
    \end{minipage}
    \hfill
    \begin{minipage}{0.32\linewidth}
        \centering
        \includegraphics[width=\linewidth,clip=True,trim=40 0 60 0]{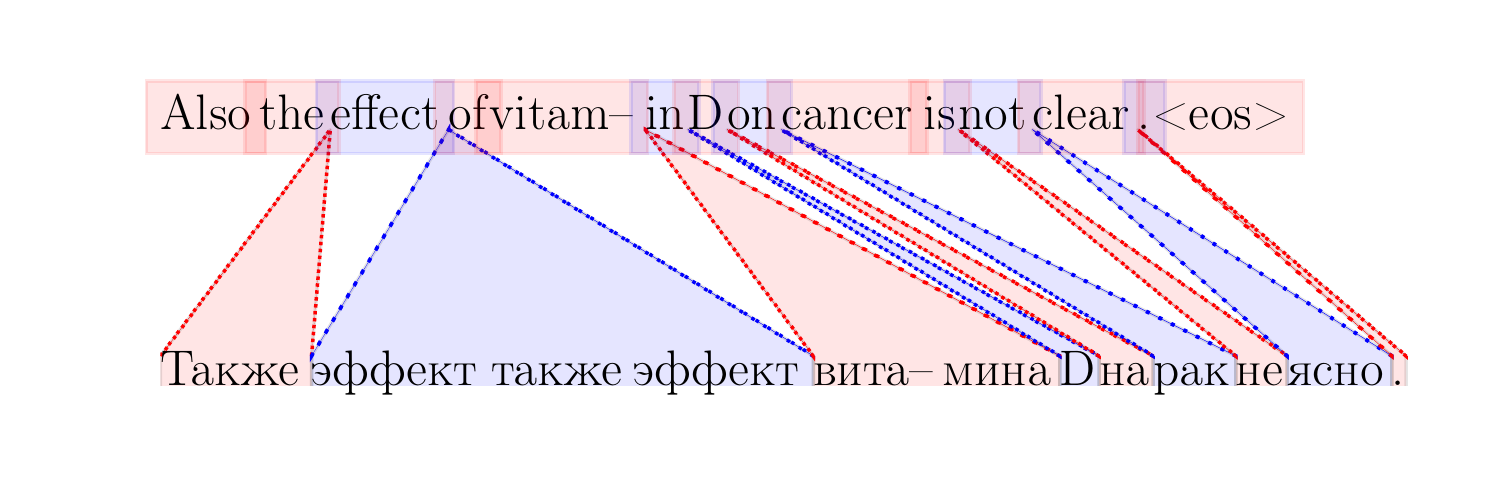}
    \end{minipage}

    \vspace{-4mm}
    \begin{minipage}{0.32\linewidth}
        \centering
        (a) {\bf Wait-If-Worse}, $\delta=1$, $s_0=2$
    \end{minipage}
    \hfill
    \begin{minipage}{0.32\linewidth}
        \centering
        (b) {\bf Wait-If-Diff}, $\delta=2$, $s_0=2$
    \end{minipage}
    \hfill
    \begin{minipage}{0.32\linewidth}
        \centering
        (c) {\bf Wait-If-Diff}, $\delta=1$, $s_0=2$
    \end{minipage}
    \caption{Example translations of ``All the effect of vitamin D on cancer is
    not clear''. The right-most plot was cut off from the right due to the space
constraint. Best viewed when zoomed digitally.}
    \label{fig:ex1}

    \vspace{-4mm}
\end{figure*}

\section{Quantitative Analysis}

\paragraph{Trade-off between Quality and Delay}

As shown in Fig.~\ref{fig:overall}, there is a clear trade-off between the
translation quality and delay. This trade-off is observed regardless of the
waiting criterion, although it is more apparent with the {\bf
Wait-If-Diff}. Out of two waiting criteria, the {\bf Wait-If-Worse} tends
to achieve a better translation quality, while it has substantially higher delay
in translation. 

Despite this general trend, we notice significant variations among the three
languages we considered. First, in the case of German, we see the trade-off
between the delay and quality is maintained in both translation directions
(En$\to$De and De$\to$En), while the delay slightly decreases when translating
to English. A similar trend is observed with Czech, except that the {\bf
Wait-If-Diff} criterion tends to improve the translation quality while
maintaining the delay.  On the other hand, the experiments with Russian show
that this trend is not universal across languages.

In the case of Russian, translation to English does not enjoy any
improvement in translation quality, as consecutive translation to English did
(compare $\bigstar$ vs. $\ding{80}$ in the right panel of
Fig.~\ref{fig:overall}.) Instead, there is a general trend of lowered delay when
translating Russian to English compared to the other way around.

We observe in all the cases that the delay decreases when the model translates
to English, compared to translating from English, with the exception of Czech
and the {\bf Wait-If-Worse} criterion. We conjecture that the richness of
morphology in all the three languages compared to English is behind this
phenomenon. Because each word in these languages has more information than a
usual English word, it becomes easier for the simultaneous greedy decoder to
generate more target symbols per source symbol. The same explanation applies
well to  the increase in delay when translating from English, as the languages
with richer morphology often require complex patterns of agreement across many
words.

\paragraph{Wait-If-Worse vs. Wait-If-Diff}

The {\bf Wait-If-Diff} criterion tends to cover wider spectra of the delay and
the translation quality. On the other hand, with the same set of decoding
parameters--$\delta$ and $s_0$--, the {\bf Wait-If-Worse} results in more
delayed translation with less variance in translation quality. In order to
further examine the difference between these two criteria, we plot the effect of
$\delta$ and $s_0$ on both the delay $\tau$ and translation quality in
Fig.~\ref{fig:criteria}. 

In Fig.~\ref{fig:criteria}, we see a stark difference between these two
criteria. This difference reveals itself when we inspect the translation quality
w.r.t. the decoding parameters (right panel of each sub-figure). The {\bf
Wait-If-Worse} criterion is clearly more sensitive to $s_0$, while the {\bf
Wait-If-Diff} one is more sensitive to $\delta$. The only exception is the case
of translating Czech to English, in which case the {\bf Wait-If-Diff} behaves
similar to the {\bf Wait-If-Worse} when $s_0\geq 5$. On the other hand, the
delay patterns w.r.t. the decoding parameters are invariant to the choice of
criterion.

\begin{figure*}[t]
    \small

    \centering
    \includegraphics[width=\linewidth,clip=True,trim=20 30 50 0]{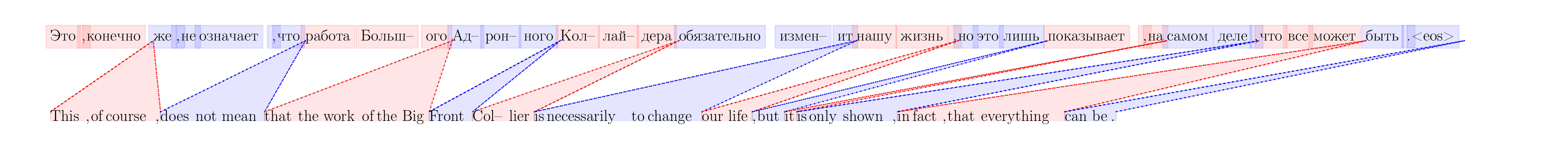}
    (a) {\bf Wait-If-Worse}

    \includegraphics[width=\linewidth,clip=True,trim=20 30 50 0]{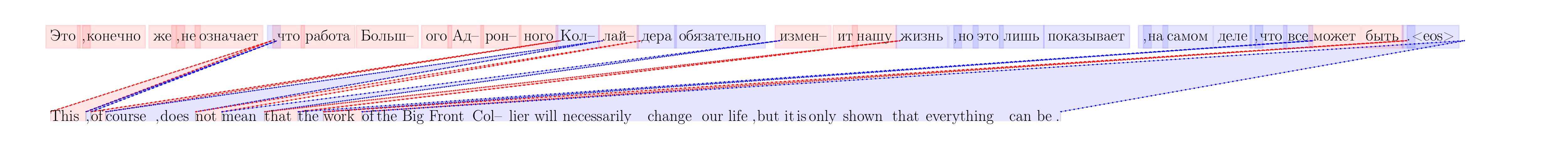}
    (b) {\bf Wait-If-Diff}

    \vspace{-2mm}
    \caption{An example of Ru$\to$En simultaneous translation. We
    observe a similar trend of conservativeness in the {\bf Wait-If-Worse}.} 
    \label{fig:ex4}

    \vspace{-4mm}
\end{figure*}

\section{Qualitative Analysis}

We define a metric of quality-to-delay ratio as $
    \text{Q2D} = \frac{\text{BLEU}}{\tilde{\tau}}$,
where $\tilde{\tau}$ is an average delay over a test corpus. We use this ratio,
which prefers a translation system with a high score and low delay, to choose
the best model for each language pair--direction. 

In Fig.~\ref{fig:example-from-en}, we present the simultaneous translation
examples by the selected models when translating from English to a target
language. We alternate between red and blue colors to indicate the
correspondence between the target symbols and their source context. The source
sentence is divided into chunks based on $s'(t)$ defined in
Sec.~\ref{sec:delay}.
We see that the {\bf Wait-If-Worse} is relatively
conservative than the {\bf Wait-If-Diff}, which was observed earlier in
Fig.~\ref{fig:overall}. 

\subsection{Ru-En Simultaneous Translation}
\label{sec:quali-ruen}

Here, we take more in-depth look at the simultaneous greedy decoding
with En-Ru as an example. 

\paragraph{Word/Phrase Repetition with Wait-If-Diff}
In Fig.~\ref{fig:ex1}, we show the Russian translations of ``All the effect of
vitamin D on cancer is not clear.'' with three different settings. The most
obvious observation we can make is again that the {\bf Wait-If-Worse} is much
more conservative than the other criterion (compare (a) vs. (b--c).) 

\begin{figure}[t]
    \centering
    \includegraphics[width=\linewidth,clip=True,trim=40 30 320 0]{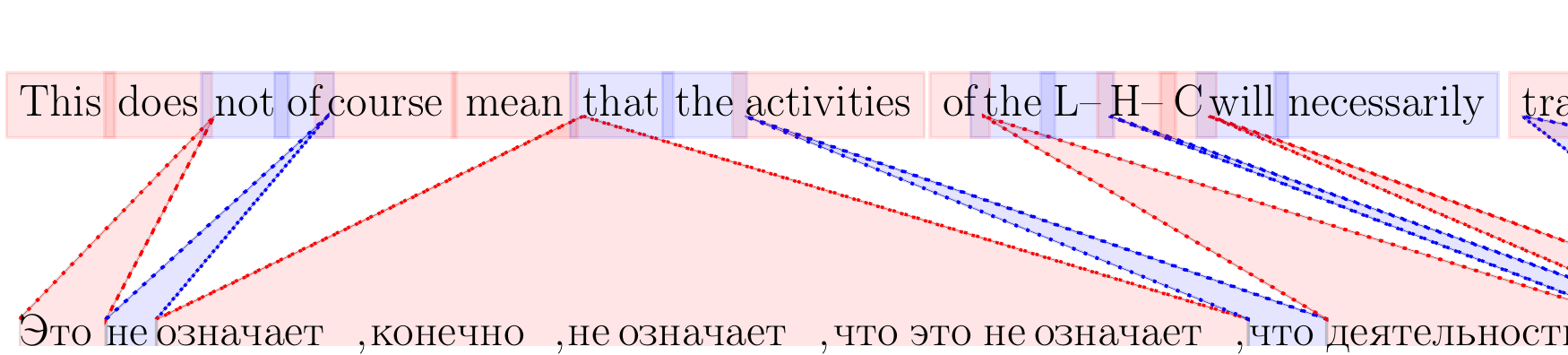}
    \caption{An example of phrase repetition when simultaneously translating
    using the {\bf Wait-If-Diff} criterion. }
    \label{fig:ex2}

    \vspace{-4mm}
\end{figure}

Another noticeable phenomenon happens with the {\bf Wait-If-Diff}, which is that
some words or phrases are repeated as being translated. For instance in
Fig.~\ref{fig:ex1}~(c), we see that ``\foreignlanguage{russian}{также эффект}'' (``also, effect'') was
repeated twice. This has been observed with other sentences. As another example,
see Fig.~\ref{fig:ex2} where ``\foreignlanguage{russian}{не означает}'' (``not means'') was
repeated three times. We conjecture that this is an inherent weakness in the
{\bf Wait-If-Diff} criterion which does not take into account the decrease in
confidence (i.e., the output conditional distribution becoming flatter) unlike
the other criterion {\bf Wait-If-Worse}. 

\paragraph{Premature Commitment}

\begin{figure}[ht]
    \centering
    \includegraphics[width=\linewidth,clip=True,trim=40 30 380 0]{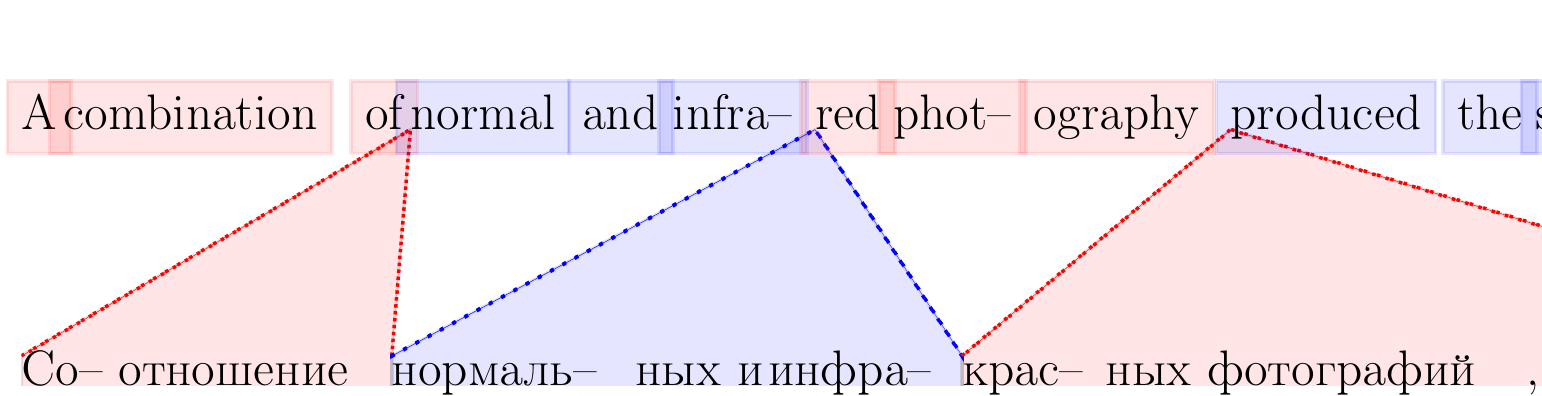}
    \caption{An example of premature commitment when simultaneously translating
    using the {\bf Wait-If-Diff} criterion. }
    \label{fig:ex3}

    \vspace{-4mm}
\end{figure}

In Russian, a preceding adjective needs to agree with the following noun. This
is unlike in English, and when translating from English to Russian, premature
commitment of adjectives may result in an inaccurate translation. For example in
Fig.~\ref{fig:ex3}, we see that the simultaneous greedy decoding committed too
early to {\it plural} adjectives (``\foreignlanguage{russian}{нормальных}'' and ``\foreignlanguage{russian}{инфракрасных}'',
``normal'' and ``infrared'') before it saw the 
noun
``photography''. The decoder translated ``photography'', which is singular, into
``\foreignlanguage{russian}{фотографи\u{и}}'' (``photos''). 

\paragraph{Russian-to-English Translation and Other Language Pairs}

In the other direction (Ru$\to$En), we observed a similar trend. For instance,
the {\bf Wait-If-Worse} tends to be more conservative than the {\bf
Wait-If-Diff} (see Fig.~\ref{fig:ex4}), while the latter more often results in
repeated phrases. Our manual inspection of other language pairs
revealed similar behaviours.

\section{Discussion and Future Research}

The quantitative analyses have revealed two major findings. First, we found that
it is indeed possible to use the neural machine translation model trained
without simultaneous translation as a goal for the purpose of simultaneous
machine translation. This was done simply by a novel simultaneous greedy
decoding algorithm described in Alg.~\ref{alg:simul_decode}. The algorithm,
which has two adjustable parameters--$s_0$ and $\delta$, allows a user to
smoothly trade off translation delay and quality, as shown in
Fig.~\ref{fig:overall}. 

The second finding is that this trade-off property depends heavily on the choice
of waiting criterion. Two criteria introduced in this paper showed markedly
opposite behaviours, where the {\bf Wait-If-Worse} was sensitive to $s_0$ while
the other, {\bf Wait-If-Diff}, was to $\delta$. We suspect that this difference
in sensitivity has led to stark qualitative differences such as the ones
discussed in Sec.~\ref{sec:quali-ruen}.

This paper presents the first work investigating the potential for
simultaneous machine translation in a recently introduced framework of neural
machine translation. Based on our findings, we anticipate further research
in the following directions. 

First, the waiting criteria proposed in this paper are both manually designed
and does not exploit rich information embedded in the hidden representation
learned by the recurrent neural networks. Information captured by the hidden
states is however difficult to extract, and we expect a trainable criterion that
takes as input the hidden state to outperform the manually designed waiting
criteria in terms of both delay and quality   

Second, all the translation models tested in this paper were not trained to do
simultaneous translation. More specifically, the decoder was exposed to a full
set of context vectors returned by the encoder only {\it after} the full source
sentence was read. A training procedure that addresses this mismatch between the
context sets seen by the decoder during training and test phases should be
designed and evaluated.


\section*{Acknowledgments}

KC thanks Facebook, Google (Google Faculty Award 2016) and NVidia (GPU Center of Excellence 2015--2016).

\bibliography{simultrans}
\bibliographystyle{emnlp2016}

\end{document}